\documentclass[11pt]{article}
\usepackage{float}

\usepackage[final]{acl}

\usepackage{times}
\usepackage{latexsym}
\usepackage{hyperref}
\usepackage[T1]{fontenc}

\usepackage[utf8]{inputenc}

\usepackage{microtype}

\usepackage{inconsolata}

\usepackage{graphicx}

\usepackage{microtype}
\usepackage{graphicx}
\usepackage{subcaption}
\usepackage{booktabs} 

\usepackage{amsmath,amssymb,amsfonts}
\usepackage{bm}
\usepackage{tikz}
\usetikzlibrary{positioning}
\usepackage{booktabs}
\usepackage{multirow}
\usepackage{url}
\usepackage{amsthm}
\usepackage{amsfonts}

\usepackage{amsthm}
\newtheorem{proposition}{Proposition}
\title{Geometry of Knowledge Allows Extending Diversity Boundaries of Large Language Models
}

\author{
  \textbf{Mateusz Bystroński\textsuperscript{1}} \And
  \textbf{Doheon Han\textsuperscript{2}} \And
  \textbf{Nitesh V. Chawla\textsuperscript{2}} \And
  \textbf{Tomasz Kajdanowicz\textsuperscript{1}}
  \AND
  \normalfont
  \textsuperscript{1}Wrocław University of Science and Technology \\
  \textsuperscript{2}University of Notre Dame \\
  \small{
    \textbf{Correspondence:}
    \href{mailto:mateusz.bystronski@pwr.edu.pl}{mateusz.bystronski@pwr.edu.pl}
  }
}

\begin{document}
\maketitle
\begin{abstract}
Starting from the hypothesis that knowledge in semantic space is organized along structured manifolds, we argue that this geometric structure renders the space explorable. By traversing it and using the resulting continuous representations to condition an LLM’s generation distribution, we can systematically expand the model’s reachable semantic range.
We introduce a framework that requires no modification of LLM parameters and operationalizes this idea by constructing a conditioning distribution from a small set of diverse anchor generations. This distribution conditions LLM's generation via an xRAG-style projector~\cite{cheng2024xragextremecontextcompression}.
Our experiments demonstrate that  this manifold-based conditioning substantially increases generative diversity, with direct benefits for enhancing divergent thinking, a core facet of creativity, in language models.

\end{abstract}

\section{Introduction}

Large language models (LLMs) have become the foundation of modern NLP systems,
yet their generative behavior exhibits a persistent limitation: despite using
stochastic decoding methods such as temperature or nucleus sampling, repeated
generations from the same prompt tend to be semantically similar. This lack of
variance constrains applications that rely on broad exploration of the semantic
space, including synthetic data generation, brainstorming, and divergent
thinking tasks.

A widely adopted strategy for increasing diversity is to manipulate the
\emph{context} presented to the model, for example through paraphrasing,
persona shifts, stylistic changes, or multi-agent discussions. Although
effective to a degree, these methods operate over a finite (or effectively
finite) set of reachable contexts. Because the conditional distribution
$p_\theta(y \mid c)$ associated with each context is known to exhibit low
variance \cite{zhang2025noveltybench}, the diversity obtainable by marginalizing over such a finite set is
inherently limited. Empirically, this manifests as rapid saturation: after only
a handful of samples, prompt- and agent-based methods cease to discover new
semantic variants.

In this work, we propose a different perspective. Instead of relying solely on
symbolic prompt manipulations, we introduce a \emph{continuous} conditioning
variable in the model's semantic space. For a given input, we derive a latent
representation and modulate the generation context via a multimodal projector,
following the xRAG mechanism \cite{cheng2024xragextremecontextcompression}. Crucially, this conditioning operates directly in
the token-embedding space of the LLM and therefore requires \emph{no
fine-tuning} of the underlying model. By exploring this continuous manifold, we
enable the model to access semantic variations that are unattainable through
prompt engineering alone.

A central challenge is determining how to sample the latent variable. We show
that classical latent models such as VAEs \cite{kingmaVAE} are ill-suited for this task due to a
topological mismatch between their unimodal latent priors and the clustered,
multi-component structure of LLM semantic representations \cite{cai2021isotropy}. Instead, we propose
an exploration-based construction: we obtain a small number of diverse anchor
responses, embed them into semantic space, and define a continuous latent
region by interpolating and perturbing these anchors. This approach naturally
supports geometric search procedures and allows semantic variation to scale
beyond the limits of prompt-based methods.

Our experimental results demonstrate that continuous semantic conditioning
substantially increases the variance of generated outputs without compromising
quality. On the \textsc{NoveltyBench} benchmark, our method uncovers new
semantic classes even at large sampling budgets while maintaining high utility.
On the Alternative Uses Test (AUT), a classical measure of divergent thinking,
latent-space exploration yields the highest originality scores across all
settings, approaching the practical upper bound of the scoring scale.

\paragraph{Contributions.}
This work makes the following contributions:
\begin{itemize}
    \item We identify a structural limitation of prompt-based and agent-based
          diversity methods, showing that their generative variance is bounded
          by the conditioning context.
    \item We introduce a plug-in latent-conditioning framework that modulates a LLM distribution through continuous exploration in semantic space,
          requiring no modification of model parameters.
    \item We provide a topological analysis explaining why
          VAE-style latent methods cannot align with the clustered geometry of
          LLM semantic activations.
    \item We empirically demonstrate substantial gains in semantic diversity and
          divergent thinking performance on \textsc{NoveltyBench} and AUT task.
\end{itemize}

\section{Related Work}
\label{sec:related}

A growing line of work has argued that contemporary LLMs suffer from
mode collapse and limited semantic variability despite stochastic
decoding. NoveltyBench~\cite{zhang2025noveltybench} introduces a
benchmark and metric suite specifically designed to assess the ability
of models to produce multiple distinct and high-quality responses to a
single prompt. Instead of relying on surface-level overlap, it clusters
outputs into abstract equivalence classes and reports diversity in
terms of the number of occupied classes and their utility. Our work
adopts this abstraction-based view of diversity and builds on
NoveltyBench as a primary evaluation environment.

Classical approaches increase variability by modifying the decoding
procedure, e.g., through temperature scaling or nucleus
sampling. These methods flatten the output distribution but do not
exploit structure across multiple generations and often exhibit a
sharp diversity–quality trade-off. Inference-time methods based on
diverse beam search~\cite{cho2016noisyparallelapproximatedecoding,li2016mutualinformationdiversedecoding,vijayakumar2017diverse,kulikov-etal-2019-importance}
and related decoding heuristics similarly operate on the token
distribution of a \emph{fixed} conditional $p_\theta(y\mid c)$:
they ensure that hypotheses in a beam differ lexically, but they do
not explicitly reason about semantic redundancy between complete
responses. Empirical comparisons with simple temperature tuning
suggest that these decoding tweaks only partially alleviate diversity
collapse and can harm quality when pushed too far~\cite{ippolito2019comparison,zhang-etal-2021-trading,peeperkorn2024temperaturecreativityparameterlarge,shurofry2024growingtailincreasingoutput}.

Recent work proposes more principled training-time mechanisms. Early
approaches encourage diversity by modifying the maximum-likelihood
objective itself: mutual-information objectives discourage generic
replies~\cite{li2016mutualinformationdiversedecoding,li-etal-2016-diversity},
unlikelihood losses penalize degenerate loops and repetitions~\cite{Welleck2020Neural},
and smoothing or reshaping the target distribution (e.g., data-dependent
Gaussian priors or explicitly diffuse targets) biases models toward
broader output distributions~\cite{Li2020Data-dependent,zhang2024forcing}. 
More recent preference-based methods encode diversity directly into the
reward or preference model: Diverse Preference Optimization and related
objectives~\cite{lanchantin2025diversepreferenceoptimization,slocum2025diverse}
and Creative Preference Optimization~\cite{ismayilzada2025creativepreferenceoptimization}
jointly optimize for quality and variety of generations. In the context
of reasoning models, online RL methods further adjust rewards or
weighting schemes to encourage exploration of diverse solution
trajectories~\cite{cui2025entropymechanismreinforcementlearning,cheng2025reasoningexplorationentropyperspective,liu2025prorl,zeng2025bstar,kirk2024understanding}. 
While effective, all of these techniques require supervised fine-tuning
or RL-style updates to the base model, which we explicitly avoid: our
goal is to increase diversity \emph{without} modifying LLM parameters. Our method is therefore complementary to these approaches.

A complementary direction explores post-hoc guidance during
generation. G2~\cite{ruan2025g2} (Guided Generation) uses an
auxiliary classifier to steer the model towards more diffuse response
distributions while maintaining task usefulness, and serves as a
strong decoding-based baseline in our experiments. Our approach is
orthogonal: rather than shaping token probabilities via an external
guidance signal, we modify the semantic \emph{conditioning} itself by
moving along a continuous manifold in embedding space.

Beyond generic methods, some work targets diversity in
application-specific formats. For instance, Holysz et. al. have explored JSON-based prompting schemes to
induce structurally diverse outputs in medical scenario \cite{holysz2025synthetic}; however,
such JSON schemas are highly task-specific and do not naturally
generalize to open-ended semantic variation. In contrast, our method
operates at the level of continuous text embeddings and applies
uniformly across tasks.

A long-standing line of research uses variational autoencoders to
endow language models with a continuous latent code controlling
generation. OPTIMUS~\cite{li2020optimus} and follow-up
work~\cite{zhang2023llamavae} train VAEs on top of large pretrained
models to organize sentences in a latent space that supports
interpolation, traversal, and conditional control. These methods, however, require optimizing the decoder. This introduces the
usual costs and risks of model fine-tuning, including potential
catastrophic forgetting of pretrained semantics. In Appendix \ref{sec:appendixb} we deep dive into this phenomenon.

xRAG~\cite{cheng2024xragextremecontextcompression} demonstrates that LLMs can be conditioned directly through dense semantic vectors injected via a multimodal projection layer. Their goal, however, is orthogonal to ours: xRAG uses embedding-based conditioning to \emph{compress external documents} for efficient RAG, keeping the model aligned to retrieved evidence. We instead generalize this mechanism to modulate the model’s \emph{internal} semantic state, treating the conditioning vector not as compressed context but as a latent variable for controlled exploration. Thus, while xRAG establishes that continuous conditioning is feasible without fine-tuning, our work leverages this capability to expand semantic variance in generation.

Our second line of evaluation concerns divergent thinking and
creativity. Recent work has begun to systematically assess language
creativity of LLMs and humans using batteries of psychological
tests. Dinu et al.\ propose an integrated creativity
suite~\cite{dinu2025testing}, including the Alternative Uses Test
(AUT), and report that strong LLMs can approach or slightly surpass
human performance under certain conditions. These works focus primarily
on measuring creativity, not on algorithmic mechanisms for increasing
it. We adopt AUT as an evaluation task and interface with an existing automatic originality scoring framework \cite{ORGANISCIAK2023101356}, which introduced a method for automated scoring that demonstrates high alignment with human annotators.

Multi-agent schemes have been explored as a way to enhance
creativity and diversity by simulating human-like group discussions.
Lu et al.\ propose \emph{LLM
Discussion}~\cite{lu2024llmdiscussionenhancingcreativity}, a three-phase multi-agent,
role-play framework which significantly improves performance on AUT
and other creativity tests compared to single-agent baselines and
simpler multi-agent setups. Subsequent surveys further document the
promise of LLM-based multi-agent systems for
creativity. Conceptually, such methods still operate within the
prompt-based paradigm: they generate a finite set of discussion
contexts and aggregate the resulting outputs. Our analysis shows that,
even with sophisticated interaction patterns, these methods remain
constrained by the low variance of $p_\theta(y \mid c)$ for each
context and by the finite size of the reachable context set. Our
experiments corroborate this: increasing the depth of LLM Discussion
yields only marginal gains before diversity saturates.

A related line of research studies interpolation-based mechanisms for
extending datasets and improving coverage of underrepresented regions
in feature space. Classical oversampling method
SMOTE~\cite{smote} generate synthetic examples by linear
interpolation between nearest neighbors and have been widely and
successfully applied to \emph{tabular data}. Deep variants such as
DeepSMOTE~\cite{deepsmote} extend this idea to
learned representation spaces, enabling interpolation in latent
embeddings learned by neural encoders in vision domain. Our approach is inspired by these techniques, but
extends the paradigm to natural language domain.

\section{Method}
\begin{figure*}
\centering
\includegraphics[width=\linewidth]{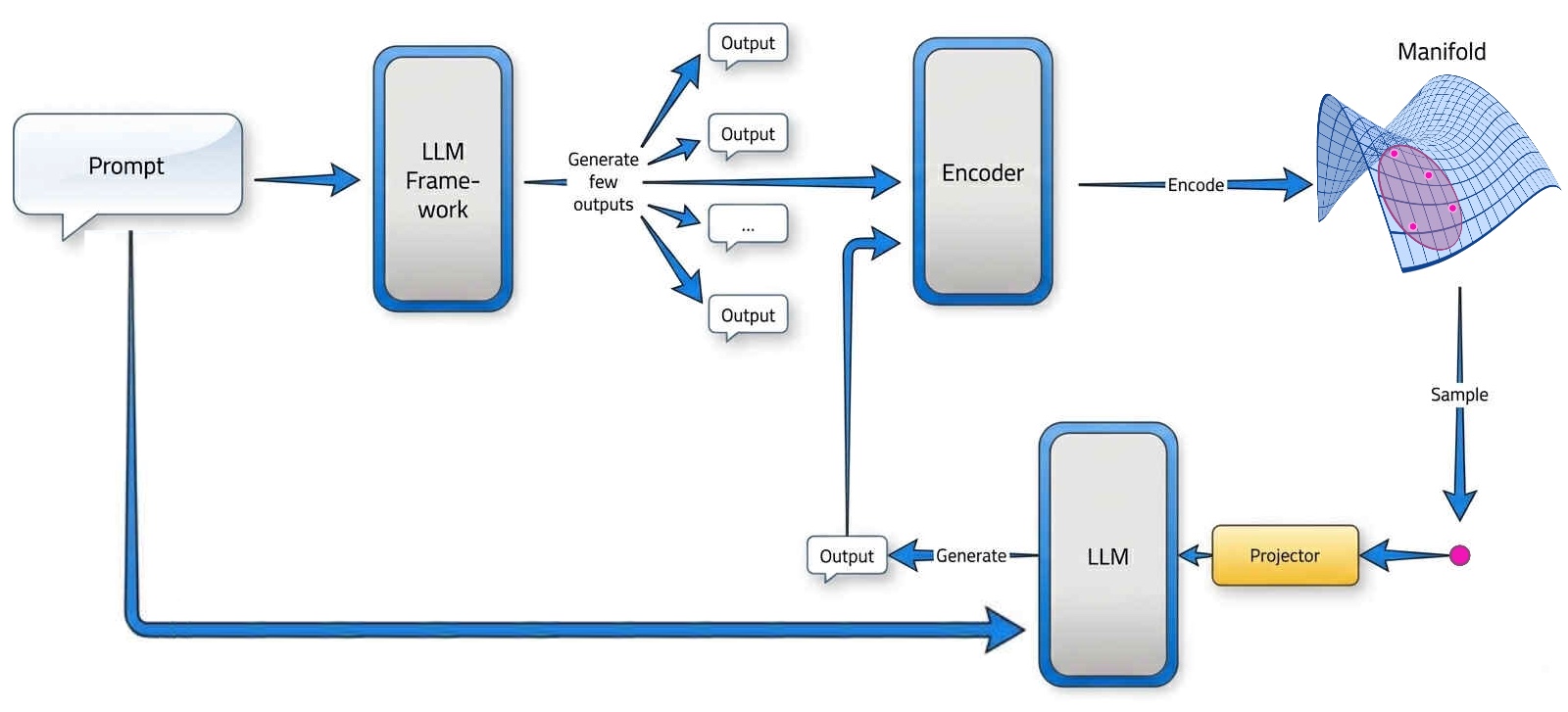}
\caption{
Given an input prompt, a base LLM first generates a small set of candidate outputs. These outputs are encoded into continuous semantic embeddings, forming a local semantic manifold. New vectors are sampled from this manifold and mapped via a xRAG projector \cite{cheng2024xragextremecontextcompression} into the LLM’s embedding space. The LLM then generates new outputs conditioned on these sampled embeddings.
}
\label{fig:method}
\end{figure*}

\subsection{Continuous semantic conditioning}

We propose approach that adds conditioning with a continuous latent variable defined in a semantic embedding space. For a given input $x$, an encoder
\begin{equation}
  e = E(x) \in \mathbb{R}^d
\end{equation}
produces a dense representation related to the task. A latent variable
\[
  z \sim q_\phi(z \mid e),\qquad z \in \mathbb{R}^d
\]
is drawn from a distribution whose support is restricted to a continuous submanifold
of the language manifold associated with \(e\).

The latent variable $z$ modulates context construction; a context is formed as
\begin{equation}
  c = g(x, z).
\end{equation}
We realize $g$ by mapping $z$ into the input token embedding space via a multimodal projector, following the approach from \cite{cheng2024xragextremecontextcompression}, in order to perform conditioning directly in the language semantic space. This allows the method to be plugged into an LLM without any fine-tuning of the language model. The LLM then generates
\begin{equation}
  Y \sim p_\theta(\cdot \mid c) = p_\theta(\cdot \mid g(x,z)).
\end{equation}
The resulting marginal distribution over outputs is
\begin{equation}
  p(y \mid x)
    = \int p_\theta\big(y \mid g(x,z)\big)\, q_\phi\big(z \mid E(x)\big)\, dz.
\end{equation}

The conditioning variable is
\begin{equation}
  z \in \mathcal{Z}_x \subseteq \mathbb{R}^d,
\end{equation}
where $\mathcal{Z}_x$ is a continuous, high-dimensional subset of the semantic space induced by the encoder.

For any feature functional $f(Y)$, the law of total variance gives
\begin{equation}
\begin{aligned}
    \operatorname{Var}[f(Y)\mid x]
    = \mathbb{E}_z\big[\operatorname{Var}[f(Y)\mid x,z]\big] + \\
      + \operatorname{Var}_z\big(\mathbb{E}[f(Y)\mid x,z]\big).
\end{aligned}
\end{equation}
In prompt based settings, that we described in Appendix \ref{sec:appendixa}, the analogue of $z$ takes values in a finite set $\mathcal{C}_x$ or is induced by weak decoding noise, which typically yields a small second term. In contrast, the latent-conditioned formulation allows $z$ to explore a continuous semantic manifold, so that
$\operatorname{Var}_z\big(\mathbb{E}[f(Y)\mid x,z]\big)$ can be substantially larger.

\subsection{Distribution of conditioning variable}

To obtain the conditioning distribution $q_\phi(z \mid e)$, we seek a method that enables semantic modulation \emph{without} fine-tuning the underlying language model, because it is computationally expensive, carries a significant risk of catastrophic forgetting, and often disrupts the delicate balance of pre-trained semantic representations.

A natural candidate is to use a variational autoencoder as the mapping \(g\), since a VAE provides a smooth latent manifold that is easy to sample from safely. However, prior work indicate that VAEs are unsuitable for this setting, as they require training with unfrozen generator \cite{li2020optimus, zhang2023llamavae}. In Appendix \ref{sec:appendixb} we explain in detail why this alignment cannot be achieved without retraining.

Instead, we propose a lightweight exploration-based approach for constructing
$q_\phi(z \mid e)$ directly in the semantic space. Although prompt-based methods
exhibit limited variance, they are nevertheless capable of producing several
distinct outputs. We treat these outputs as \emph{anchor points} and embed them
into the semantic space, forming a discrete set
\[
  A_x = \{e_1, \ldots, e_m\},
\]
which serves as a basis for defining the latent region $\mathcal{Z}_x$.

In our experiments we restrict exploration to interpolation-based families.
This choice is motivated by the use of an encoder trained with contrastive
objectives, which tends to organize semantic classes into approximately convex
clusters. As a result, interpolation between semantic anchors produces stable
and meaningful latent variations.
Sampling from $q_\phi(z \mid e)$ is defined explicitly through anchor selection
and interpolation:
\[
  (i,j) \sim \pi, 
  \qquad 
  \lambda \sim \rho,
\]
The latent variable is then constructed as
\[
  z = (1-\lambda)e_i + \lambda e_j.
\]
That produces a continuous latent region $\mathcal{Z}_x$
spanned by the anchor set $A_x$.

We also find considerable potential in using random perturbations around anchors and, more broadly, meta-heuristic search techniques (e.g., evolutionary exploration or gradient-free manifold traversal). These strategies allow $q_\phi(z \mid e)$ to cover richer and more diverse semantic regions without modifying the LLM itself. Exploring such families of search-based latent distributions constitutes a promising direction for future work.

\section{Experiments}
Following setup from \cite{cheng2024xragextremecontextcompression}, in all experiments we use \textbf{Mistral-7B-Instruct} \cite{jiang2023mistral7b} as the underlying language model responsible for generating textual outputs. Semantic representations are obtained using the \textbf{Mistral SRF embedding model} \cite{SFRAIResearch2024}, which serves as the encoder producing continuous embedding vectors used by our exploration distribution \( q_{\phi}(z \mid e) \).
To instantiate the conditioning function $g(x,z)$, we employ the
\emph{projector} introduced in \cite{cheng2024xragextremecontextcompression}, which maps latent
vectors $z$ into the token-embedding space of the LLM.

\subsection{Generation diversity}

We evaluated the ability of our method to generate diverse responses on the
\textsc{NoveltyBench} benchmark on the \textit{curated} dataset, which quantifies output diversity using the
\textit{Distinct} metric. Following the benchmark specification%
~\cite{zhang2025noveltybench}, \textit{Distinct} is defined as the number
of abstract equivalence classes obtained by clustering semantically similar
generations. Formally, for a set of $k$ generations,
\[
\mathrm{distinct}_k := \bigl|\{c_i : i = 1, \dots, k\}\bigr|,
\]
where $c_i$ denotes the functional equivalence class assigned to the $i$-th
generation. Importantly, these classes are not derived from raw embedding
similarity alone: the benchmark employs a trained classifier to identify
conceptual usage categories, ensuring that the metric captures semantic rather
than surface-level diversity.

Diversity can be trivially increased by producing responses that are
uninformative or misaligned with the task. To account for this, we additionally
report the \textit{Utility} metric, which evaluates not only whether a generation
belongs to a new equivalence class, but also whether it remains meaningful and
useful within the task specification. NoveltyBench defines cumulative utility as
\[
\mathrm{utility}_k := \frac{1 - p}{1 - p^k}
\sum_{i=1}^{k} p^{\,i-1}
\cdot 
\mathbf{1}\Bigl[c_i \neq c_j \ \forall j < i\Bigr]
\cdot u_i,
\]
where $p$ is the user patience parameter (set to $0.8$ in the benchmark),
$\mathbf{1}[\cdot]$ indicates whether the $i$-th output introduces a new
equivalence class, and $u_i$ denotes the utility score assigned to that
generation.

We evaluate both metrics across multiple generation budgets (10, 15, 20, 25, 30 samples). It is particularly important for diversity to increase with larger sampling budgets, especially in applications such as synthetic data generation, where the marginal gains from additional samples directly translate into broader and more representative coverage of the underlying semantic space.

As initialization points for defining the anchor set $A_x$, we used a first generations produced by G2. We selected 30\% of the initial outputs for the 10-20 sample budgets and 20\% for the 25 and 30 budgets.

To induce high semantic variability within this region, we employed an aggressive exploration strategy. The choose is motivated by ablation studies. W sampled  $\lambda$ coefficient from intervals:
\[
  \lambda \sim U([6, 10] \cup [-6, -10]).
\]

However, we observed that latent conditioning can occasionally introduce noise and structural drift in the generated outputs, for instance, producing responses that differ from the expected format (e.g., returning a paragraph with seven sentences instead of five). To mitigate this, we introduced an additional alignment step: after generating a candidate response, we submitted it back to the model with an explicit instruction to realign the output to the task specified by prompt.

\subsection{Results}

\begin{table*}[t]
  \centering
  \small
  \begin{tabular}{llccccc}
  \toprule
  & & \multicolumn{5}{c}{NoveltyBench (k generations)} \\
  \cmidrule(lr){3-7}
  Method & Metric 
         & $k\!=\!10$ & $k\!=\!15$ & $k\!=\!20$ & $k\!=\!25$ & $k\!=\!30$ \\
  \midrule
  \multirow{3}{*}{Standard} 
    & Distinct (\#)      & 4.37 & 4.64 & 5.43 & 6.20 & 6.79 \\
    & Distinct-\% (\%)    & 43.7 & 30.9 & 27.2 & 24.8 & 22.6 \\
    & Utility            & 3.62 & 0.84 & 0.82 & 0.78 & 0.79 \\
  \midrule
  \multirow{3}{*}{In-context} 
    & Distinct (\#)      & \textbf{7.13} & \underline{9.35} & \underline{11.70} & \underline{12.68} & 13.31 \\
    & Distinct-\% (\%)    & \textbf{71.3} & \underline{62.3} & \underline{58.5} & \underline{50.7} & 44.4 \\
    & Utility            & 2.97 & 2.81 & 2.71 & 2.76 & 2.71 \\
  \midrule
  \multirow{3}{*}{G2} 
    & Distinct (\#)      & 6.21 & 8.29 & 10.27 & 12.04 & \underline{13.60} \\
    & Distinct-\% (\%)    & 62.1 & 55.3 & 51.3 & 48.2 & \underline{45.3} \\
    & Utility            & \underline{4.52} & \underline{4.48} & \underline{4.31} & \underline{4.33} & \underline{4.33} \\
  \midrule
  \multirow{3}{*}{Ours (G2 seeds)} 
    & Distinct (\#)      & \underline{7.10} & \textbf{10.11} & \textbf{12.31} & \textbf{14.12} & \textbf{16.65} \\
    & Distinct-\% (\%)    & \underline{71.0} & \textbf{67.4} & \textbf{61.6} & \textbf{56.5} & \textbf{55.5} \\
    & Utility            & \textbf{4.78} & \textbf{4.79} & \textbf{4.56} & \textbf{4.55} & \textbf{4.59} \\
  \bottomrule
  \end{tabular}
  \caption{
  Results for different methods across various generation counts (k). Distinct represents the mean number of distinct partitions, Distinct-\% is the percentage of distinct generations (distinct/k $\times$ 100\%), and Utility is the mean discounted utility from the scoring pipeline.
  }
  \end{table*}

\noindent
\noindent The results indicate that our latent-conditioning approach reliably expands the variance of generated outputs across all sampling budgets. As the number of generations increases, the method continues to uncover new semantic variants rather than collapsing into repetitive patterns. Importantly, this increase in variance is not accompanied by any degradation in usefulness: utility remains consistently high, demonstrating that the latent-modulated contexts remain aligned with the underlying task objectives. Taken together, the findings provide strong empirical evidence that continuous conditioning in semantic space is an effective and stable mechanism for enhancing generative diversity while preserving output quality.

\subsection{Impact of Diverse Generation on Divergent Thinking Capabilities}

We examine how increased generation diversity translates into improved divergent thinking capabilities.
We laveraged psychological Alternative Uses Test (AUT), a classical psychological assessment of divergent thinking and creative potential~\cite{lu2024llmdiscussionenhancingcreativity, Silvia2008DivergentThinking, dinu2025testing}. Participants in an AUT task are asked to propose unusual and non-obvious uses for everyday objects. Responses are traditionally evaluated for originality, flexibility, and fluency. We focus on core metric, originality \cite{lu2024llmdiscussionenhancingcreativity}.

We used the originality scoring framework from \cite{ORGANISCIAK2023101356}, which provides automated originality ratings (from 1 to 5) aligned with human-labeled AUT datasets, using model \textbf{ocsai-4o}, which is said by authors to be good for English Alternate Uses scoring. Following the creativity literature~\cite{Silvia2008DivergentThinking, dinu2025testing}, which recommends focusing on only a few ideas, we report:
\begin{itemize}
    \item \textbf{Top--1 originality}: the most original idea,
    \item \textbf{Top--2 originality}: the mean of the two most original ideas,
    \item \textbf{Top--3 originality}: the mean of the three most original ideas.
\end{itemize}

We laveraged dataset from \cite{lu2024llmdiscussionenhancingcreativity}

For each AUT prompt, we used the output of the multi-turn \textbf{LLM discussion} as the anchor set.  
Previous work reports that this discussion-based method does not scale \cite{lu2024llmdiscussionenhancingcreativity}; we confirmed this in our own setup by running discussions of varying lengths.  
The best-performing discussion depth was selected, and its output served as both our baseline and our anchor points.

We evaluated \textsc{G2}  using the first generations from the LLM discussion as contextual seeds.  
Our proposed method was likewise evaluated using the LLM discussion outputs as anchors.

For both methods, we ran 500 generations per approach.

We did not apply any alignment step during the AUT experiment. Since the task focuses purely on the originality of semantic content rather than consistency of style or structure, additional stylization mechanisms were unnecessary and were therefore omitted.

\subsection{Results}

\begin{table}[ht]
\centering
\small
\begin{tabular}{lccc}
\toprule
\textbf{Method} & \textbf{Top--1} & \textbf{Top--2} & \textbf{Top--3} \\
\midrule
LLM Discussion (1 round) & 4.17 & 4.06 & 3.95 \\
LLM Discussion (3 rounds) & 4.57 & 4.52 & 4.49 \\
LLM Discussion (5 rounds) & 4.58 & 4.55 & 4.50 \\
LLM Discussion (7 rounds) & 4.58 & 4.56 & 4.53 \\
\midrule
\textbf{G2} 
& \underline{4.93} & \underline{4.92} & \underline{4.90} \\
\midrule
\textbf{Ours} 
& \textbf{4.99} & \textbf{4.98} & \textbf{4.95} \\
\bottomrule
\end{tabular}
\caption{
Comparison of AUT originality scores across discussion-based baselines,
\textsc{G2}, and our latent-space exploration method.  
Scores are reported as Top--1, Top--2, and Top--3 originality.
}
\label{tab:aut-originality-expanded}
\end{table}

\begin{figure}[ht]
\centering
\includegraphics[width=\linewidth]{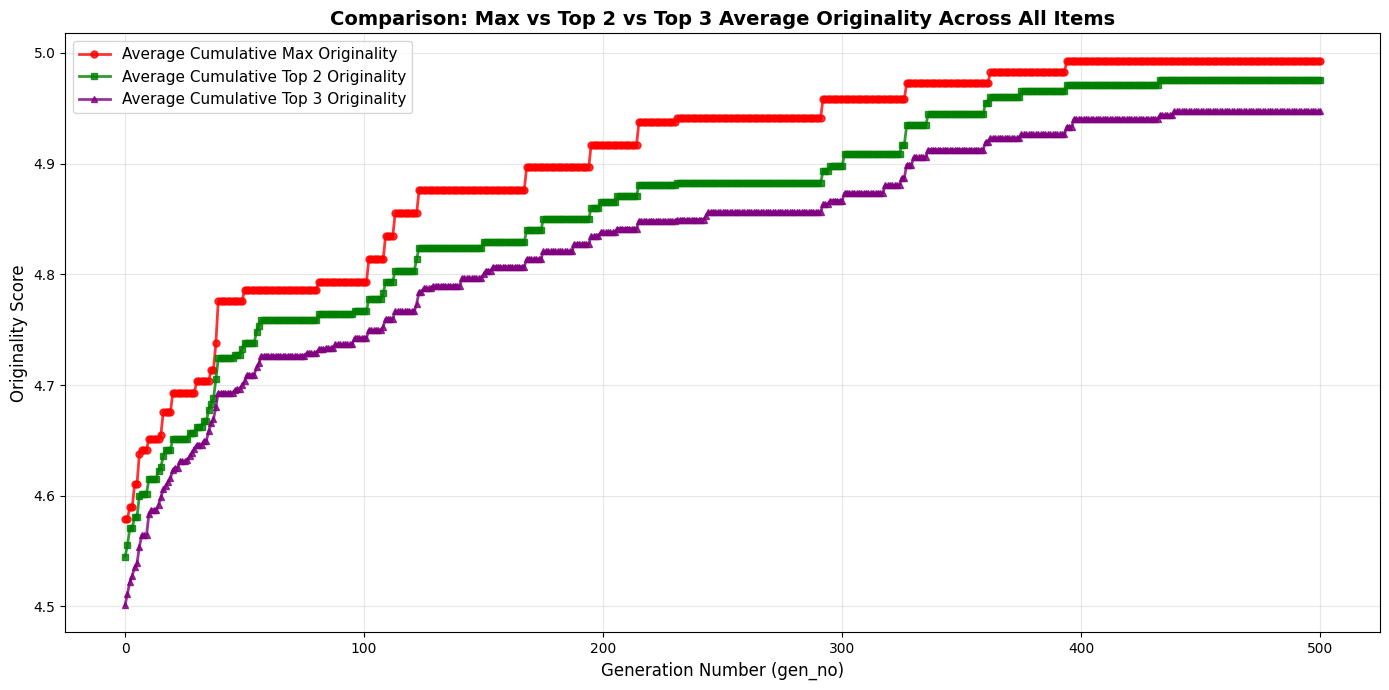}
\caption{
Cumulative originality curves for our latent-space exploration method.
As more latent samples are drawn, the Top--1, Top--2, and Top--3 originality
scores steadily increase.
}
\label{fig:ours-originality-curve}
\end{figure}

The results highlight three key observations. First, increasing the
depth of multi-agent LLM discussion yields only marginal gains: Top-1
plateaus after a few rounds, confirming that this method does not
scale. Second, expanding the diversity of generations has a direct and measurable
impact on creativity. As illustrated at Figure \ref{fig:ours-originality-curve}, Top--1, Top--2, and Top--3 scores steadily improve as more latent
samples are drawn, indicating that broader exploration of the semantic space
translates into consistently more original ideas. Finally, our latent-space
method achieves the strongest originality across all evaluation settings,
reaching a Top--1 score of 4.99. This value is extremely close to the practical
upper bound of the AUT scale - 5, effectively demonstrating that our method pushes the model’s creative capacity to the limits.

\section{Ablation Studies}

\subsection{Lambda Ablation Study}

\begin{figure*}[t]
\centering
\includegraphics[width=\linewidth]{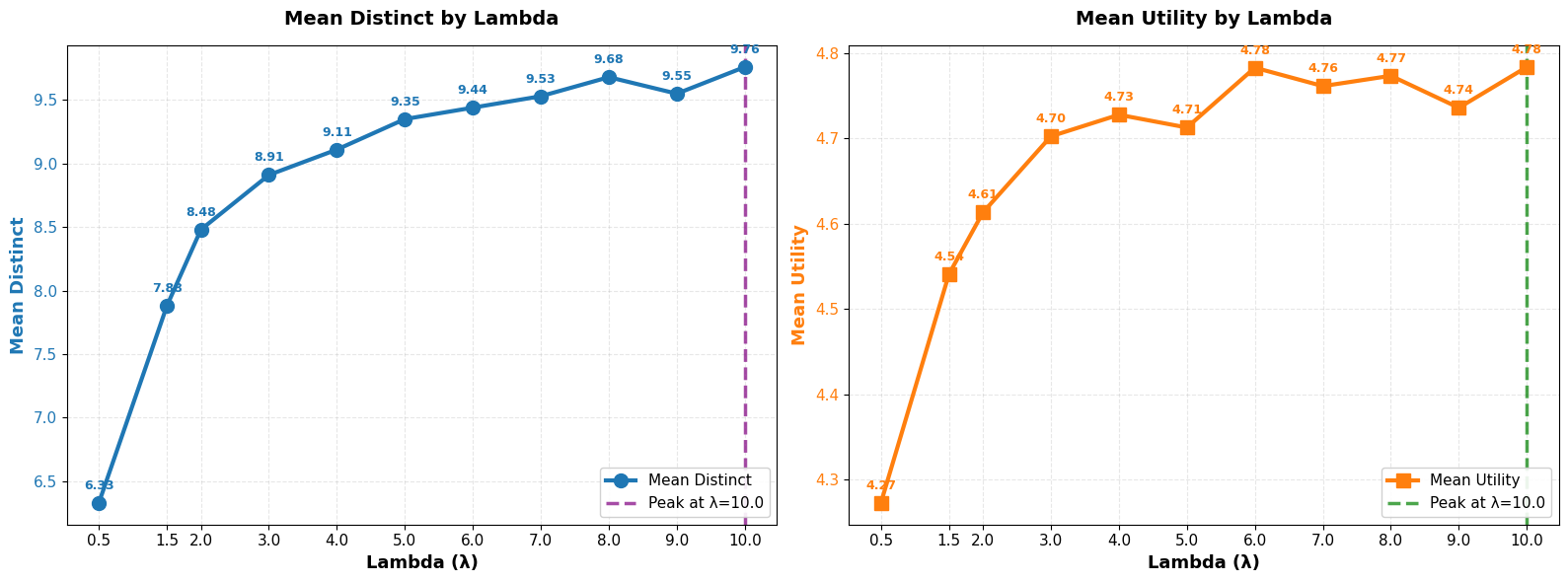}
\caption{
Ablation over \(\lambda\). Small values keep the latent variable inside the
anchor cluster, yielding low diversity; larger values explore broader semantic
regions, improving diversity without harming quality.}
\label{fig:lambda-ablation}
\end{figure*}

Figure~\ref{fig:lambda-ablation} shows how the interpolation coefficient
\(\lambda\) influences diversity and utility. We observed low
Distinct score at \(\lambda = 0.5\), indicating that prompt-based anchor
generations form a tight semantic cluster: staying close to them yields only
minor variations. Increasing \(\lambda\) moves the latent variable outside this
region, leading to a sharp rise in diversity while maintaining high utility.
Performance peaks around \(\lambda = 10\), showing that broader interpolation
unlocks genuinely new semantic direction.
These results directly motivate the sampling range used in our main diversity
experiments.

\subsection{Ablation: Dependence on Anchor Seeds}

\begin{table*}[t]
  \centering
  \small
  \begin{tabular}{llccccc}
  \toprule
  & & \multicolumn{5}{c}{NoveltyBench (k generations)} \\
  \cmidrule(lr){3-7}
  Anchors source & Metric 
         & $k\!=\!10$ & $k\!=\!15$ & $k\!=\!20$ & $k\!=\!25$ & $k\!=\!30$ \\
  \midrule
  \multirow{3}{*}{In-context} 
    & Distinct (\#)      & 7.1 & 10.23 & 13.24 & 15.03 & 17.85 \\
    & Distinct-\% (\%)    & 71.0 & 68.2 & 66.2 & 60.1 & 59.5\\
    & Mean Scores        & 3.32 & 3.07 & 2.81 & 2.66 & 2.54\\
  \midrule
  \multirow{3}{*}{G2} 
    & Distinct (\#)      & 7.1 & 10.11 & 12.31 & 14.12 & 16.65 \\
    & Distinct-\% (\%)    & 71.0 & 67.4 & 61.6 & 56.5 & 55.5 \\
    & Mean Scores        & 3.93 & 3.58 & 3.24 & 2.89 & 2.82 \\
  \bottomrule
  \end{tabular}
  \caption{
  Results for our method with In-context and G2 anchors across various generation counts (k). Distinct represents the mean number of distinct partitions, Distinct-\% is the percentage of distinct generations (distinct/k $\times$ 100\%), and Mean Scores is the mean of all generation scores.
  }
  \label{tab:seed-ablation}
  \end{table*}

In this experiment we isolate the effect of anchor quality on latent-space
conditioning. Instead of the utility metric, we report the mean score of all
generations, as utility is highly dependent on first generations.

The results in Table~\ref{tab:seed-ablation} show clear dependence on the chosen
anchors. Using in-context anchors yields higher Distinct values but noticeably
lower mean scores, reflecting the weakness of in-context prompting from earlier experiment:
it produces diverse but low-quality seeds, and our method inherits this trade-off.
Conversely, G2 anchors provide stronger and more coherent starting points,
leading to consistently higher mean scores while achieving lower distinct value.

\section{Discussion}

Our experimental results demonstrate that the proposed method introduces substantially greater variation in generated outputs, and that this variance translates into responses that are both more diverse and fully comparable in quality to those produced by baseline approaches.

The lack of scalability observed in LLM discussion further confirms our theoretical considerations: because the underlying variance of the conditional distribution is intrinsically low, the model is unable to discover regions of higher originality within the semantic space. Subsequent perturbations introduced by additional discussion rounds remain too small to meaningfully expand this variance, leading to rapid saturation.

A key insight is that the latent variable we introduce is not a randomness, it is a variable that has its own context that directly influences the model’s response, analogous to the role of retrieved evidence in RAG systems, and this context can be explored geometrically. This makes it possible to apply a wide range of heuristics and metaheuristics to text representations. For example, evolutionary crossbreeding can be naturally expressed as a linear combination of text embeddings.

The AUT experiment illustrates this particularly clearly: we started from outputs generated by a complex agent-based method, and used these responses as anchors and further optimized them through latent-space exploration. In essence, the way we did it in experiment was as an evolutionary strategy, where whole population breeds together.

Taken together, these findings open a new perspective on NLP: tasks traditionally limited by the symbolic and contextual nature of natural language can now be addressed using classical methods from computer science, enabled by the continuous and geometrically structured semantic space

\section{Limitations}

We used ChatGPT as a writing assistant to improve the clarity and readability of the manuscript.
All scientific content, experimental design, analysis, and conclusions were developed and verified by the authors.

Despite strong results, our approach has several limitations.

First, it does not explicitly detect low-quality or out-of-distribution generations. Apart from a heuristic realignment step in some experiments, there is no built-in mechanism for hallucination or factuality control.

Second, the exploration distribution \(q_\phi(z \mid e)\) is simplistic. We use fixed-range linear interpolations with a scalar \(\lambda\), ignoring the local geometry of the embedding space (e.g., density, cluster structure). The same \(\lambda\)-range is applied across prompts and tasks, which can lead to under- or over-exploration.

Finally, latent moves are restricted to linear combinations of anchor embeddings and depend heavily on the chosen anchors and embedding stack. Weak anchors yield diverse but low-quality generations, and we have not yet evaluated robustness across different LLMs, encoders, or domains.

\bibliography{custom}

\appendix

\section{Analysis of Prompt-Based Methods}
\label{sec:appendixa}

A large language model (LLM) is viewed as a conditional distribution
\begin{equation}
  p_\theta(y \mid c),
\end{equation}
where $y$ denotes the generated output sequence, $c$ denotes the input context (system prompt, user prompt, in-context examples, retrieved documents, etc.), $\theta$ are fixed model parameters.

For a given task input $x$ (e.g., a user instruction), a deterministic context constructor
\begin{equation}
  c = g(x)
\end{equation}
is applied, and the model generates
\begin{equation}
  Y \sim p_\theta(\cdot \mid g(x)).
\end{equation}
All stochasticity arises from the sampling procedure used to decode from $p_\theta$ (for example, temperature sampling or nucleus sampling), while the initial condition is fully specified by the single context $c$, and it is shown that the nature of LLMs makes this distribution have low variance \cite{zhang2025noveltybench}.

\subsection{Prompt transformations}

Many existing approaches to improve diversity operate by applying a finite collection of prompt transformations. Let
\begin{equation}
  c_k = T_k\big(g(x)\big), \qquad k = 1,\dots,K,
\end{equation}
denote the transformed contexts obtained from a fixed input $x$. Methods such as paraphrasing the instruction, switching personas, or adding stylistic variations can all be modeled as choosing one element from the finite set
\begin{equation}
  \mathcal{C}_x = \{c_1,\dots,c_K\}.
\end{equation}

Generation proceeds by first sampling a context
\begin{equation}
  C \sim \mu_x,
\end{equation}
where $\mu_x$ is a distribution supported on $\mathcal{C}_x$, and then sampling
\begin{equation}
  Y \sim p_\theta(\cdot \mid C).
\end{equation}

Let $D(\cdot)$ be any functional measuring diversity of a distribution over outputs (e.g., entropy or an expected semantic distance). The induced marginal distribution of $Y$ for a fixed $x$ can be written as
\begin{equation}
  p(y \mid x)
    = \sum_{k=1}^K \mu_x(c_k)\, p_\theta(y \mid c_k),
\end{equation}
and therefore
\begin{equation}
\begin{aligned}
    D\big(p(Y \mid x)\big)
    = D\!\left( \sum_{k=1}^K \mu_x(c_k)\, p_\theta(\cdot \mid c_k) \right) \leq \\
    \leq f\!\left(K, \{p_\theta(\cdot \mid c_k)\}_{k=1}^K\right),
\end{aligned}
\end{equation}
for some problem-dependent upper bound $f$. In particular, the attainable diversity is structurally controlled by the finite set $\mathcal{C}_x$ and the conditional distributions $\{p_\theta(\cdot \mid c_k)\}$ that are known to have low variance.

\subsection{Iterative in-context prompting}

Iterative schemes such as chain-of-thought prompting and self-consistency can be modeled as a sequence of contexts
\begin{equation}
  c^{(1)} = g(x), \qquad
  y_1 \sim p_\theta(\cdot \mid c^{(1)}),
\end{equation}
\begin{equation}
  c^{(2)} = h\big(c^{(1)}, y_1\big), \qquad
  y_2 \sim p_\theta(\cdot \mid c^{(2)}),
\end{equation}
and, in general,
\begin{equation}
  c^{(t)} = H_t\big(x, y_{1:t-1}\big), \qquad
  y_t \sim p_\theta(\cdot \mid c^{(t)}),
\end{equation}
for $t = 1,\dots,T$. The final answer is either $y_T$ or some aggregation of $(y_1,\dots,y_T)$.

This entire procedure can be abstracted as a stochastic operator
\begin{equation}
  Y = F_\theta(x, \varepsilon),
\end{equation}
where $\varepsilon$ collects all randomness introduced by decoding at each step. For any feature functional $f$ of the output, the variance is given by
\begin{equation}
  \operatorname{Var}\big[f(Y)\mid x\big]
    = \operatorname{Var}_\varepsilon \big[f(F_\theta(x,\varepsilon))\big].
\end{equation}
Empirically, the conditional distributions $p_\theta(\cdot \mid c^{(t)})$ often exhibit low variance \cite{zhang2025noveltybench}. As a consequence, this variance $\operatorname{Var}_\varepsilon$ tends to remain modest, reflecting that the system remains effectively driven by the initial condition $x$ together with a relatively weak stochastic perturbation $\varepsilon$.
\subsection{Agents Will Not Help}

Multi-agent methods instantiate several LLM agents that exchange messages before producing a final answer. Let
\begin{equation}
  c_i^{(1)} = g_i(x), \quad
  y_i^{(1)} \sim p_\theta(\cdot \mid c_i^{(1)}), \quad i = 1,\dots,M,
\end{equation}
denote the first round of messages. Subsequent rounds update each agent's context, for example,
\begin{equation}
  c_i^{(2)} = h_i\big(x, y_{1:M}^{(1)}\big), \quad
  y_i^{(2)} \sim p_\theta(\cdot \mid c_i^{(2)}),
\end{equation}
and so on, until an aggregation module produces a final output $Y$ from all intermediate messages.

Again, this can be written in compact form as
\begin{equation}
  Y = G_\theta(x, \varepsilon),
\end{equation}
where $\varepsilon$ encompasses all decoding randomness across all agents and rounds. For any feature functional $f$ of the output,
\begin{equation}
  \operatorname{Var}\big[f(Y)\mid x\big]
    = \operatorname{Var}_\varepsilon \big[f(G_\theta(x,\varepsilon))\big],
\end{equation}
and, due to the limited stochasticity of the underlying conditional distributions, this variance tends to remain modest.

\subsection{Conclusion}

Collecting the above, prompt-based and multi-agent methods can be viewed as
operating within the model class
\begin{equation}
  \mathcal{M}_{\mathrm{prompt}}
    = \big\{ p_\theta(y \mid c) : c \in \mathcal{C}_x \big\},
    \qquad |\mathcal{C}_x| < \infty,
\end{equation}
where $\mathcal{C}_x$ contains all contexts reachable from a fixed input $x$ under
the given protocol and decoding scheme. Since these methods marginalize only over
a finite (or effectively finite) set of reachable contexts, the diversity of the
resulting output distribution is fundamentally constrained.

This limitation is made explicit by the law of total variance. For any feature
functional $f(Y)$,
\begin{equation}
\begin{aligned}
    \operatorname{Var}[f(Y)\mid x]
    = \mathbb{E}_{c \sim \mu_x}\!\left[
        \operatorname{Var}\big[f(Y)\mid x, c\big]
      \right] + \\
      + 
      \operatorname{Var}_{c \sim \mu_x}\!\left(
        \mathbb{E}\big[f(Y)\mid x, c\big]
      \right).
\end{aligned}
\end{equation}
First term i s shown to be modest \cite{zhang2025noveltybench}, and because $c$ ranges only over the finite set $\mathcal{C}_x$, the second term is
structurally bounded by $\mathcal{C}_x$ itself and cannot increase beyond the
variance induced by choosing among these finitely many contexts.

\begin{figure*}
\centering
\includegraphics[width=\linewidth]{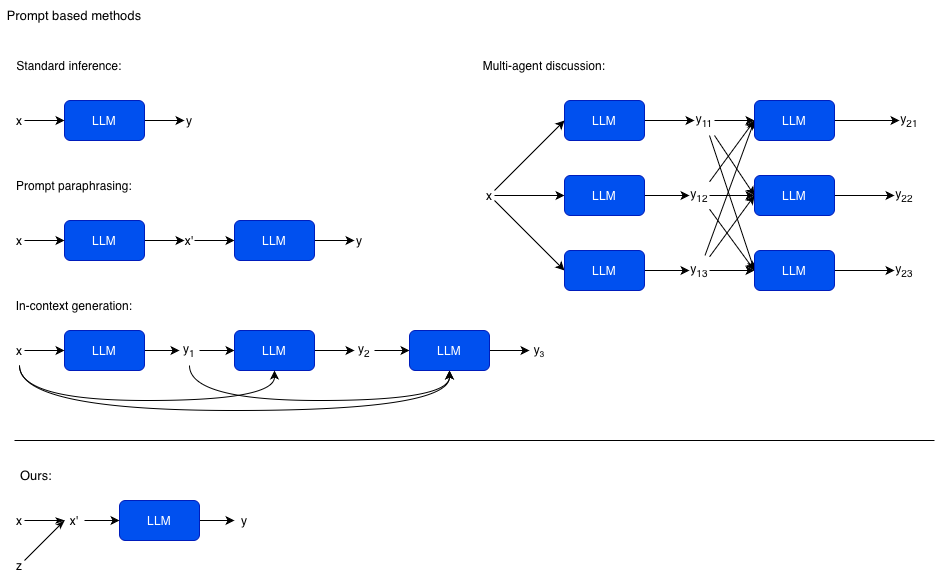}
\caption{
Graphical comparition of prompt based methods with our approach.
}
\label{fig:comparition}
\end{figure*}

Figure \ref{fig:comparition} shows graphical comparition of prompt based methods with our approach.

\section{VAE Projector Leads to OOD Risk}
\label{sec:appendixb}
Let \(Z\subset\mathbb{R}^{d}\) denote the support of the VAE latent distribution used for sampling, and let
\(H\subset\mathbb{R}^{d_h}\) denote the support of the LLM's semantic decoder--input
activations (e.g., first-layer token embeddings).  
We model \emph{semantic clusters} in the LLM as the path--connected components of
high-density regions in \(H\).  

Let \(p_Z\) be the latent density (typically Gaussian), and let \(p_H\) denote the empirical
density of decoder activations.  
For a threshold \(\epsilon>0\), define the super-level sets
\[
  Z_\epsilon = \{\,z\in Z : p_Z(z)\ge \epsilon\,\}, 
\]
\[
  H_\epsilon = \{\,h\in H : p_H(h)\ge \epsilon\,\}.
\]
Because \(p_Z\) is intentionally smooth and unimodal, the set \(Z_\epsilon\) is a
\emph{single connected component}.  
In contrast, the decoder-side super-level set decomposes into $l$ multiple semantic
clusters [literatura]:
\[
  H_\epsilon = \bigsqcup_{j=1}^{\ell} D_j.
\] 

\subsection{Splitting Implies Valley Traversal}

Let a continuous decoder-conditioning map \(f : Z \to H\) represent the process of
feeding a sampled latent vector into the LLM's semantic space (e.g.,
via a multimodal projector).  
If \(Z_\epsilon\) is connected but \(H_\epsilon\) decomposes into multiple components,
then no continuous \(f\) can map the single latent region into multiple semantic
clusters without traversing the low-density valleys between them.

\begin{proposition}[VAE Splitting Implies Semantic Valley Traversal]
\label{prop:vae_split_valley}
Assume the “ground-truth” semantic assignment would require
\[
  Z_\epsilon \;\longrightarrow\; D_{j_1} \;\cup\; D_{j_2},
  \qquad 
  D_{j_1}\cap D_{j_2}=\emptyset.
\]
Let
\[
  V_\tau = \{\,h\in H : p_H(h)<\tau\,\}
\]
be the low-density \emph{valley set} separating decoder clusters, for some
\(\tau < \epsilon\).
If a continuous \(f\) satisfies
\[
  f(Z_\epsilon) \cap D_{j_1}\neq\emptyset
  \quad\text{and}\quad
  f(Z_\epsilon) \cap D_{j_2}\neq\emptyset,
\]
then necessarily
\[
  f(Z_\epsilon) \cap V_\tau \;\neq\; \emptyset.
\]
Thus any continuous splitting of the single latent component into multiple decoder
semantic islands must traverse the valley between them.
\end{proposition}

\begin{proof}[Sketch]
Since \(Z_\epsilon\) is path-connected and \(f\) is continuous, the image
\(f(Z_\epsilon)\) is also path-connected.  
A path connecting a point in \(D_{j_1}\) to one in \(D_{j_2}\) must leave
\(D_{j_1}\cup D_{j_2}\) and enter their complement, which is contained in
the valley \(V_\tau\).  
Hence \(f(Z_\epsilon)\) intersects \(V_\tau\).
\end{proof}

\subsection{Out-of-Distribution Risk in VAE Sampling}

Because a VAE imposes a \emph{single, connected} latent region from which sampling must
cover the entire space, it cannot align its latent topology with the inherently
clustered structure of LLM semantic space \cite{cai2021isotropy}.  
Any attempt to map a single VAE latent component onto multiple semantic clusters
forces the image of latent samples to pass through low-density regions
\(
  V_\tau
\). In order to change semantic space on initial layers of LLM, fine tuning is necessary.
\end{document}